# Dueling Deep Q Network for Highway Decision Making in Autonomous Vehicles: A Case Study

Teng Liu*, Xingyu Mu*, Xiaolin Tang*, Bing Huang*, Hong Wang**, Dongpu Cao***

*College of Automotive Engineering, Chongqing University, Chongqing, 400044, China
(Tel: +86 13638219888; e-mail: tengliu17@gmail.com,
20162361@cqu.edu.cn, tangxl0923@cqu.edu.cn, 20162364@cqu.edu.cn,).
**School of Vehicle and Mobility, Tsinghua University, Beijing, China
(e-mail: hong_wang@tsinghua.edu.cn)
***Department of Mechanical and Mechatronics Engineering, University of Waterloo, N2L 3G1, Canada
(e-mail: dongpu_cao@uwaterloo.ca)

**Abstract:** This work optimizes the highway decision making strategy of autonomous vehicles by using deep reinforcement learning (DRL). First, the highway driving environment is built, wherein the ego vehicle, surrounding vehicles, and road lanes are included. Then, the overtaking decision-making problem of the automated vehicle is formulated as an optimal control problem. Then relevant control actions, state variables, and optimization objectives are elaborated. Finally, the deep Q-network is applied to derive the intelligent driving policies for the ego vehicle. Simulation results reveal that the ego vehicle could safely and efficiently accomplish the driving task after learning and training.

## 1. INTRODUCTION

Owing to the rapid development of artificial intelligence, autonomous driving has become an essential technology all over the world [1-4]. The consumers, vehicle manufactures, policymakers, and governments are paying much attention to the critical techniques of autonomous vehicles (AVs) [5]. Many benefits are capable of being achieved by automated vehicles, such as crashes reduction, remission of traffic congestion, improvement of driving enjoyment, and the promotion of efficiency and safety [6-7]. However, to realize full automation, the remarkable processes are necessary for the modules of perception, decision-making, planning, and control in AVs [8].

Decision-making strategy in autonomous driving represents determining the lateral and longitudinal control actions at each time instance [9]. This policy is hugely affected by other traffic participants, such as pedestrians, surrounding vehicles, traffic lights, and driving environments. These players usually contain uncertainties, and thus the decision-making module needs to predict their behaviors and make the right decisions [10]. In general, the driving environments for AVs can be cast into two types, highway and urban ones. Urban driving environments are more complex than the highway ones because they include more participants and uncertainties [11].

Inspired by the great success of deep learning (DL) and reinforcement learning (RL) in many research areas, more and more literature is applying these approaches to address the critical problems in AVs. For example, Duan et al. [12] studied a hierarchical RL control framework to optimize the decision-making strategy of self-driving vehicles. The high-level focused on the maneuver selection and the low-level controlled the lateral and longitudinal motions. Ref. [13] investigated the Bayesian RL method, wherein an ensemble of neural network (NN) with additional randomized prior functions are included. The relevant decision-making policy is proved to be available for the uncertainty of decisions. Furthermore, the authors in [14] and [15] combined the long short-term memory network with RL approaches to construct the predictive decision-making strategy for automated cars on the highway. The predicted behaviors of surrounding vehicles are regarded as expert knowledge to search optimal control actions. However, the traditional RL techniques are not able to resolve the complicated driving environments because the ample search space of states and actions would restrict the computational efficiency.

In this work, a dueling deep reinforcement learning (DRL)-based control architecture is proposed to address the highway overtaking problem for AV. First, the highway driving environments are introduced, wherein the number of lanes and the number of surrounding vehicles are generalized. It derived the decision-making policy is easily transferred to other driving situations. Then, the utilized DRL technique is determined, including the formulation of state variables, control actions, and optimization objectives. Finally, a series of simulation experiments are conducted to prove the effectiveness of the presented highway decision-making policy. The obtained driving strategy is also applied in the real-world driving data to illuminate its advantages in efficiency and safety.

There are three perspectives of contributions are included in this article:1) A DRL-based decision-making structure is presented for the highway decision-making problem of automated vehicles; 2) the particular DRL method, named dueling deep Q networks (DDQN) is designed to acquire the overtaking polciy for lateral and longitudinal motions; 3) a series evaluation experiments are constructed to demonstrate the effectiveness of the proposed decision-making policy after mature learning and training.

The following organization of this work is designed as follows: Section II gives the modeling of highway decision-making problem; the DRL algorithm and solving process of the built problem are described in Section III. Section IV executes a comprehensive analysis of the obtained simulation results, and finally Section V concludes this article.

## 2. HIGHWAY OVERTAKING PROBLEM FORMULATION

This section discusses the studied highway decision-making problem. The driving environments of the ego vehicle is founded. The speed, acceleration, and position are taken as the state variables and control actions in this problem. The information of surrounding is assumed to be known to indicate the uncertainties. Furthermore, the optimal control goal is designed to imitate real-world driving conditions.

### 2.1 Driving Environments on Highway

The highway is a relatively undemanding driving scenario for the implementation of autonomous driving. Several leading automobile companies, such as Ford, Tesla, Waymo, Audi, General Motors and so on have tested their automated products on the highway [16]. However, the supervised decision-making controls are necessary under human drivers. It motivates scientific researchers to develop a more advanced management controller for the decision-making system.

In this work, the highway decision-making problem for AVs is transferred into an optimization control problem and solved by the DRL technique. The decision-making policy indicates determining the lateral and longitudinal motions. As described in Fig. 1, a particular three-lanes highway scenario is considered. The yellow car is referred to as the ego vehicles, and the green ones are the surrounding vehicles. The objective of the ego vehicle is driving through the highway scenario as fast as possible without causing collisions.

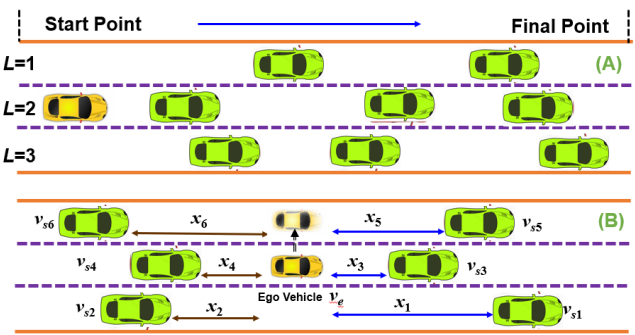

Fig. 1. Three-lanes highway driving scenario for decision-making problem.

For easy understanding, the lanes from left to right are named as $L$ = 1, 2, and 3. Assuming the number of surrounding vehicles is 3*$N$, it implies that there are $N$ cars on each lane. All the vehicles are running from left to right, and the surrounding vehicles operate lane-changing behaviors randomly, which causes the uncertainties for the driving situations. For the ego vehicle, the starting and ending points are pre-defined. Hence, the ego vehicle should drive from the starting to ending point with an efficient strategy.

### 2.2 Vehicle Kinematics

This subsection discusses the vehicle dynamics of the simulation vehicles. Since the lateral and longitudinal directions are both considered, they are mimicked by the Cartesian coordinate system, and the positions are computed as follows [17]:

$$\begin{cases} \dot{x} = v\cos(\varphi + \beta) \\ \dot{y} = v\sin(\varphi + \beta) \end{cases} \quad (1)$$

where $(x, y)$ is the geographic position of the vehicle, $v$ is the vehicle velocity, $\varphi$ and $\beta$ are the heading and slip angles. Then, the acceleration and these two angels are further represented as:

$$\begin{cases} \dot{v} = a \\ \dot{\varphi} = \dfrac{v}{l}\sin\beta \\ \beta = \tan^{-1}(1/2\tan\delta) \end{cases} \quad (2)$$

where $l$ is the wheel tread and $\delta$ is the front wheel angle. After calculating the speed and acceleration of each vehicle, the relative distance and velocity between the ego and nearby vehicles are depicted as:

$$\begin{cases} \Delta x_i = |x_i - x_e| \\ \Delta v_i = |v_i - v_e| \end{cases}, i = 1, 2, ... \quad (3)$$

where the subscript $i$ and $e$ indicate variables of surrounding and ego vehicles. In this work, the relative distance and velocity are settled as the state variables. Thus, the ego vehicle should maximize the state variable by action choice.

To control the ego vehicle on the lateral and longitudinal derections, five control actions are considered. They are formulated as the following expression:

$$a_e = \begin{cases} Left\_lane, & change\ to\ left\ lane \\ Maintain, & keep\ speed\ and\ lane \\ Right\_lane, & change\ to\ left\ lane \\ Acceleration, & driver\ fast \\ Deceleration, & drive\ slowly \end{cases} \quad (4)$$

At each step, the ego vehicle would choose only one control action to achieve the goal. This objective propels the ego vehicle to keep driving efficiency and safety in mind. It means the ego vehicle would not crash other vehicles and pass through the highway scenario as soon as possible.

The default configuration of the ego vehicle and surrounding vehicles are the same. The length and width of the vehicle are 5.0 m and 2.0 m, respectively. The maximum reachable speed is 40 m/s. Furthermore, the initial speed of the ego is randomly selected from [23, 25] m/s. The simulation environment of this work is settled in Python 3.7 [18], the visualization window is shown in Fig. 2. The particular parameters of the DRL methods are given in the next section.

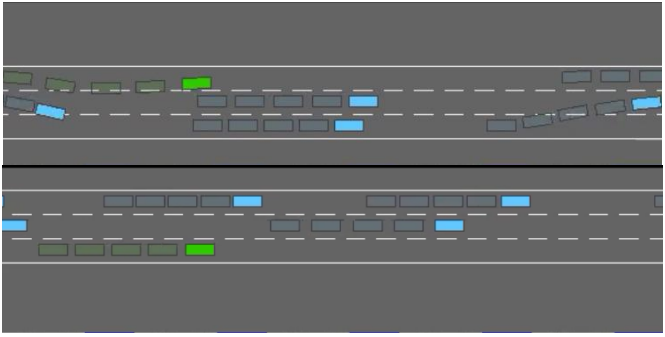

Fig. 2. Sample window of the simulation environment in Python.

## 3. RL FRAMEWORK AND DDQN ALGORITHM

This section introduces the RL framework, including the crucial arguments, such as state variable, control action, reward, transition model. Then, the utilized dueling deep Q networks (DDQN) is discussed in detail. The advantage of this algorithm is analyzed theoretically.

### 2.1 RL Framework

Machine learning is composed of supervised, unsupervised and reinforcement learning (RL) approaches. The significant characteristic of RL realizes self-evaluation and self-improvement via the interaction between the agent and the environment. RL algorithms have many classifications, such as model-based and model-free, policy-based, and value-based, Monte-Carlo and temporal-difference. Many methods have been proven to be effective in multiple research fields [19, 20].

In the interaction of RL, the current control action would affect not only the immediate reward but also the future ones. Hence, the Markov decision processes (MDPs) are usually applied to mimic this interactive process. A tuple (*S*, *A*, *P*, *R*, *β*) is always used to represent the MDP, wherein *S* and *A* are the set of state variables and control actions. *P* is the transition model of the state in the environment, and *R* is the reward function to indicate the good or bad of action choice. Finally, *β* is a called discount factor to maintain the balance of immediate and future rewards.

In the RL, the agent aims to seek an optimal control sequence $\pi$ to maximize (or minimize) the expected discounted reward, which is defined as follow:

$$R_t = \sum_t^\infty \beta^t r_t \quad (5)$$

where t is the time step, *r* is the instantaneous reward. In the value-based RL algorithms, two value functions are usually computed regarding the state *s* or state-action pair (*s*, *a*), respectively:

$$\begin{cases} Q^\pi(s,a) = E[R_t \mid s_t, a_t, \pi] \\ V^\pi(s) = E_{a \sim \pi}[Q^\pi(s,a)] \end{cases} \quad (6)$$

The value function V implies how good it is to be in a special state *s*. And the state-action function Q measures the value of selecting *a* fixed action when in this state. To compute the state-action function (Q function), a recursive expression is formulated with policy $\pi$:

$$Q^\pi(s,a) = E[r + \beta E_{a' \sim \pi}[Q^\pi(s',a')] \mid s,a,\pi] \quad (7)$$

where *a'* and *s'* are the next action and state. To obtain the optimal control action at each time instance, the optimal Q function is written as the Bellman equation:

$$Q^*(s,a) = E[r + \beta \max[Q(s',a')] \mid s,a] \quad (8)$$

In the Deep Q network, a neural network is employed to approximate the Q function, which is described as $Q(s, a; \theta)$. The loos function between the actual and approximate Q function is formulated as follows:

$$\begin{cases} L(\theta) = E[(y^{DQN} - Q(s,a;\theta))] \\ y^{DQN} = r + \beta \max_{a'} Q(s',a';\theta^-) \end{cases} \quad (9)$$

where $\theta^-$ us the argument of a fixed and separate target network used to update the parameters of neural network. Finally, to acquire the optimal approximate Q function in DRL algorithms, the loss function is calculated by the gradient descent method as:

$$\nabla L(\theta) = E[(y^{DQN} - Q(s,a;\theta))\nabla_\theta Q(s,a;\theta)] \quad (10)$$

In the highway driving scenario, the current action choice may not cause collision, but it may influence the performance of the following selection. Hence, the dueling network is applied to estimate the advantage of action choice. To bring

this insight into reality, an advantage function is introduced [21] as:

$$A^{\pi}(s,a)=Q^{\pi}(s,a)-V^{\pi}(s) \quad (11)$$

It means in the DDQN algorithm, one more neural network is added to approximate the advantage function (A function). The Q function in the DDQN algorithm is them rewritten as:

$$Q^{\pi}(s,a;\theta,\alpha)=V^{\pi}(s;\theta,\alpha)+(A^{\pi}(s,a;\theta,\alpha)-\max_{a'} A^{\pi}(s,a';\theta,\alpha)) \quad (12)$$

By doing this, the agent can estimate the value and advantage of the control action choice at each step, and it would result in better convergence and control performance.

In this work, the reward function for the ego vehicle is related to the collision conditions and the vehicle speed as follows:

$$\begin{cases} r=-10, & \text{collision happen} \\ r=-(v_e-v_{e,\max})^2, & \text{collision not happen} \end{cases} \quad (12)$$

In each episode, the ε-greedy policy is applied to choose the control action [22]. For specification, the discount factor $\beta$ is 0.8, and the $\varepsilon$ decreases from 1 to 0.05 with exponential decay of time constant 6000. The total training episodes are 2000, and in each episode, the maximum score is 20. In the next section, different simulations are executed to evaluate the presented DDQN-based decision-making policy on the highway.

## 4. RESULTS AND DISCUSSION

This section would analyze the control performance of the proposed highway decision-making strategy for AVs. First, the training process with a significant number of episodes is presented. The state variables are depicted as simulation results to prove the availability of the proposed algorithm. Then, the characteristics of normal DQN and dueling DQN are compared. The relevant results indicate the DDQN could realize better performance in this work.

*4.1 Training Process of Highway Decision-Making*

In the training process of DDQN-based highway decision-making policy, the number of episodes is settled as 2000, the initial position is at the lane 2 (*L*=2), the starting speed is randomly chosen from [23, 25] m/s and the maximum reward is defined as 20. The variation of total reward in each episode is depicted as Fig. 3. It is evident that the total reward increases with the episodes, and after about 500 trials, the ego vehicle could reach the maximum score. And in the episodes from 1000 to 2000, the ego vehicle could reach the ending point in most cases.

Since the reward includes two parts, the collision conditions, and speed limitation, Fig. 4 shows the traveling distance (Length in y label) and the average speed of the ego vehicle. It should be noticed that if the collision happens, the current episode will interrupt directly, and thus the traveling distance of ego vehicle will not be long enough in this case. From Fig. 4, it can be discerned that the collision does not happen in the later period and the vehicle could reach the destination. Since the average speed will also affect the traveling distance, the traveling distance would change in the later period. Furthermore, to get more rewards, the ego vehicle learns to boost its speed and run more efficiently.

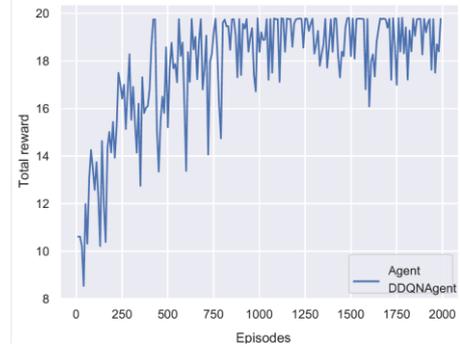

Fig. 3. Total reward of training process for DDQN-based decision-making policy.

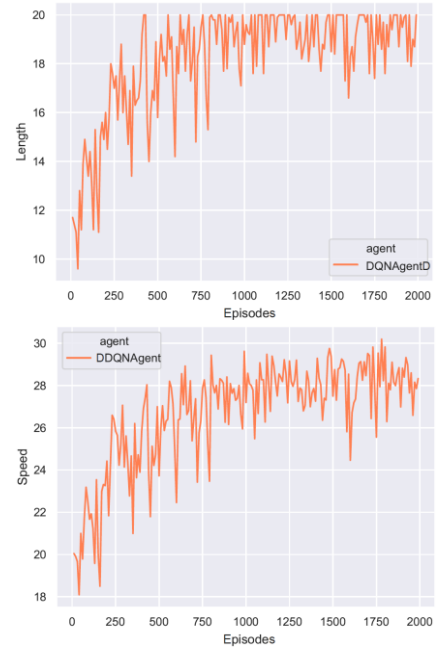

Fig. 4. Traveling distance and average speed of ego vehicle in training course.

*4.2 Comparison of DQN and DDQN*

To display the advantage of advantage network (in (11)) in this work, the conventional DQN and DDQN are compared first. The mean error of Q function between these two algorithms is shown in Fig. 5. As the Q function influences the selection of control actions, the DDQN is able to obtain the optimal in a more efficient way. The downtrend of these two trajectories means the ego vehicle in these two cases obtained more and more knowledge about the driving environments along with the learning process. Hence, the ego vehicle could drive efficiently and safely. It also implies that the agent in DDQN could be more familiar with the environment with the same number of episodes.

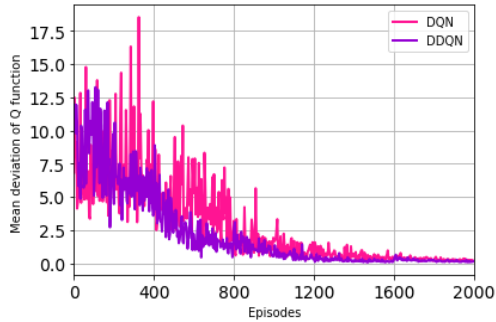

Fig. 5. Mean error of Q function in two DRL methods: DQN and DDQN.

Fig. 6 describes the accumulated reward of these two approaches. The uptrend of these two curves also indicates the ego vehicle is learning to choose the control actions to obtain more rewards. Furthermore, the cumulative reward in DDQN is always larger than the DQN algorithm, and thus the ego vehicle in DDQN could achieve the optimal control policy faster. This is caused by the advantage network, which helps the ego vehicle to quantify the potential values of the selected control actions.

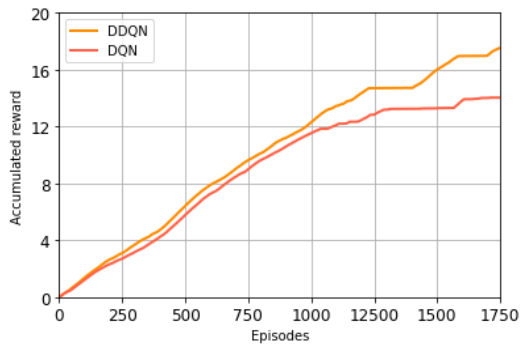

Fig. 6. Cumulative rewards with episodes in DQN and DDQN.

## 6. CONCLUSIONS

This paper discusses the DDQN-based overtaking highway policy for automated vehicles. The dueling network is suitable for the decision-making process of AVs on the highway. The related simulation shows the DDQN could obtain the optimal control policy after learning and training. Furthermore, the comparative analysis between DQN and DDQN indicates the proposed method is appropriate to get a safe and efficient driving policy.